\documentclass[twocolumn]{article}

\usepackage[round]{natbib}

\usepackage{algorithm}
\usepackage{algorithmic}
\usepackage{amssymb}
\usepackage{amsmath}
\usepackage{mathrsfs}
\usepackage{amsthm}
\usepackage{dsfont}
\usepackage{multirow}
\usepackage{graphicx}
\usepackage{paralist}
\usepackage{booktabs}
\usepackage{titlesec} 

\titlespacing*{\section}
{0pt}{*0.8}{*0.8} 
\titlespacing*{\subsection}
{0pt}{*0.5}{*0.5}
\titlespacing*{\subsubsection}
{0pt}{*0.3}{*0.3}

\usepackage[blocks]{authblk}
\renewcommand\Affilfont{\small} 

\begin{document}

\title{AMUSE: Adaptive Model Updating using a Simulated Environment}

\author[1]{Louis Chislett}
\author[2]{Catalina A. Vallejos}
\author[3]{Timothy I. Cannings}
\author[4]{James Liley}
\date{}

\makeatletter
\renewcommand\AB@affilsepx{\\ \protect\Affilfont}
\makeatother

\affil[1]{\textbf{Corresponding Email:} louis.chislett@ed.ac.uk}
\affil[1,2]{MRC Human Genetics Unit, University of Edinburgh, Crewe Rd S, Edinburgh, EH4 2XU, United Kingdom.}
\affil[3]{School of Mathematics, University of Edinburgh, University of Edinburgh, EH9 3FD, United Kingdom}
\affil[4]{Department of Mathematical Sciences, Durham University, Stockton.
Rd, Durham, DH1 3LE, United Kingdom}
\affil[2]{The Alan Turing Institute, 96 Euston Rd, London, NW1 2DB, United Kingdom.}

\twocolumn[
  \begin{@twocolumnfalse}
    \maketitle
    \begin{abstract}
        Prediction models frequently face the challenge of concept drift, in which the underlying data distribution changes over time, weakening performance. Examples can include models which predict loan default, or those used in healthcare contexts. Typical management strategies involve regular model updates or updates triggered by concept drift detection. However, these simple policies do not necessarily balance the cost of model updating with improved classifier performance. We present AMUSE (Adaptive Model Updating using a Simulated Environment), a novel method leveraging reinforcement learning trained within a simulated data generating environment, to determine update timings for classifiers. The optimal updating policy depends on the current data generating process and ongoing drift process. Our key idea is that we can train an arbitrarily complex model updating policy by creating a training environment in which possible episodes of drift are simulated by a parametric model, which represents expectations of possible drift patterns. As a result, AMUSE proactively recommends updates based on estimated performance improvements, learning a policy that balances maintaining model performance with minimizing update costs. Empirical results confirm the effectiveness of AMUSE in simulated data.
    \end{abstract}
  \end{@twocolumnfalse}
]

\section*{Introduction}

Classification models are statistical or machine learning approaches to predict a binary outcome $Y$ based on its relationship to a set of observed covariates or features $X$~\citep{Hastie2001TheLearning}. For example, QRISK3~\citep{Hippisley-Cox2017DevelopmentStudy} predicts 10-year risk of cardiovascular disease based on health and demographic information. Traditionally, classifiers are trained once on some historical data, and then applied for predicting on some new, unseen data. However, in the real-world the system which underpins the relationship between $X$ and $Y$ often evolves, changing the distribution of the data over time. Such changes are often referred to as `concept drift', and occur in unforeseen ways~\citep{Widmer1996}. The performance of the classifier may deteriorate in response to this changing relationship~\citep{Gama2014AAdaptation}, prompting the need to update the classifier on more recent data, to improve performance.

Concept drift has a diverse range of causes. It can be abrupt~\citep[e.g.~changes in 
hospitalization patterns due to the COVID-19 pandemic;][]{Duckworth2021UsingCOVID-19} or gradual (e.g.~increasing disease prevalence in an aging population). Models may simply be updated at arbitrary time points, often triggered by data availability or logistical reasons. For instance, the QRISK model is currently on its third iteration~\citep{Hippisley-Cox2017DevelopmentStudy}.
Instead, several methods have been developed to detect concept drift~\citep{Lu2018LearningReview}. Such methods can signal alerts to prompt `updating' of the classifier in some way~\citep{Lu2018LearningReview}. Examples include the popular Drift Detection Method (DDM) \citep{Gama2004LearningDetection} and other related approaches \citep[e.g.][]{Frias-Blanco2015OnlineBounds, Baena-Garcia2006EarlyMethod}. 

The methods mentioned above prompt model updating alerts solely based on \emph{observed} concept drift. This paper provides a proof-of-concept, in which we combine observed concept drift with the \emph{expected} improvement in performance of a classifier (upon re-training) to derive a model updating procedure that better balances accuracy gains with the costs incurred from model updating (e.g.~resources to re-train a complex classifier, or collecting new data). AMUSE (Adaptive Model Updating using a Simulated Environment) creates a virtual environment in which episodes of concept drift are simulated. By representing the sequential model updating procedure as a Markov Decision Process (MDP)~\citep{Puterman1990MarkovProcesses}, AMUSE uses a Reinforcement Learning (RL) algorithm~\citep{daoun2022} to learn an initial model updating policy prior to future, potentially drifted, real world data arriving. This initial policy is then iteratively improved upon observing its performance on the real world, potentially drifted data. We give heuristic reasoning for our approach alongside simulated results demonstrating comparative long-term predictive performance with fewer updates than existing methods.

\section*{Related work}

Several drift detection methods are available in the existing literature. For example, the Drift Detection Method (DDM) uses the error rate of the classifier to construct a hypothesis test for detecting drift \citep{Gama2004LearningDetection}. This motivated the development of other error rate based methods such as the Hoeffding's inequality based Drift Detection Method (HDDM) ~\citep{Frias-Blanco2015OnlineBounds}, the Early Drift Detection Method (EDDM)~\citep{Baena-Garcia2006EarlyMethod}, Learning with Local Drift Detection (LLDD)~\citep{Gama2006LearningDetection} and Fuzzy Windowing Drift Detection Method (FW-DDM)~\citep{Liu2017FuzzyAdaptation} and the ensemble-based approach by \citet{Minku2011DDD:Drift}. Other approaches for concept drift detection include FLORA \citep{Widmer1996} and the method by \citet{Klinkenberg2000DetectingMachines}, both designed to select a relevant data subset to be used when updating a model.

This paper is not the first to consider the model updating procedure as a MDP, however it is not common. Model updating as an MDP was first proposed by \citet{Liebman2018ALearning}, where a policy was learned through Approximate Value Iteration. This method involved creating the state space by discretizing core distributional properties of the data and the model. However, unlike in~\cite{Liebman2018ALearning}, our method involves the creation a simulating environment where we learn a policy prior to observing drifted data.

\section*{Problem Setting}

Let $X \in \mathscr{X}$ be  covariate values from an unspecified space, and $Y \in \{0,1\}$ be binary outcome values. Concept drift can be defined as any change in the joint distribution of $(X,Y)$ over time~\citep{Gama2014AAdaptation, Lu2016ATechnique, Losing2016KNNDrift}. Here we
focus on changes in the conditional distribution of $(Y|X)$~\citep[sometimes called `real concept drift' or `posterior drift',][]{Gama2014AAdaptation}, assuming that the marginal distribution of $X$ does not change. 
For example, changes in operational procedures within a hospital may change the distribution of an outcome $Y$ (e.g.~whether a patient has an emergency hospital admission) with respect to historic cases with similar covariates $X$~\citep[e.g.~age and sex;][]{Duckworth2021UsingCOVID-19}.

At a high level, our aim is to determine optimal update times of a classifier in the presence of concept drift in an online fashion (sequentially as we observe data). In particular, we wish to maintain a high accuracy while minimising the number of updates recommended, and thus minimising any costs associated with updating. Our setting is motivated by the following example of a so called clinical prediction model:

\emph{A classifier is trained to predict a binary health outcome (e.g.~type 2 diabetes incidence), based on demographic information as well as health information collected when a patient interacts with the health system. Data arrives at discrete time intervals (e.g.~monthly) and it is with this data that a model could be updated, for which we mean a complete re-training of the classifier. In this setting, re-training a new classifier carries with it a significant `cost'. For example re-training a classifier requires analyst time, or a model must undergo significant evaluation in order to be deemed safe for patients. We wish to maximise the cumulative utility of the classifier over its life-span, given the costs associated with updates. Here, utility refers to a metric of performance of the model at each time step, for instance, the correct classification rate.}

\subsection*{Notation and Aim}
 
Let $T \in \mathbb{N}$ be a positive, finite time horizon. We will refer to $T$ as the `model life-span', the total period of time in which a classifier will be used, regardless of how many updates that are made to it. For example, a model may have a life-span of 10 years with 5 updates during that time. We write $[T] = \{1, \ldots, T\}$ and suppose at each time $t \in [T]$, we observe a dataset $\Delta_t$ consisting of $n_t$ independent samples from the joint distribution of $(X,Y)$ at time $t$. 

We assume we are able to fit, on any such dataset $\Delta$, a data-dependent classifier $C$, which takes inputs $x\in \mathscr{X}$ and returns  $C(x; \Delta) \in [0,1]$, the predicted probability of $(Y=1|\{X =x\})$. We will write $C_t \equiv C(\cdot, \Delta_t)$ for ease of notation. At time $t=1$ we fit the classifier $C_1$.

At each time $t$ we define an action $a_t \in \Omega = \{0,1\}$ which decides whether we update the model ($a_t=1$) or keep the existing model ($a_t=0$). We denote $u(t)\in [t]$ as the `most recent update' at time $t$, that is $u(t) = \max\{t':a_{t'} =1,t' \leq t\}$ so that the model used at time $t$ is $C_{u(t)}$. Thus we have at each time $t$:
\begin{equation}
u(t) = 
\begin{cases} 
   u(t-1) &\text{ if } a_t=0 \\
   t &\text{ if } a_t=1. 
\end{cases} \label{eq:utdef}
\end{equation}

We always set $a_1 = 1$ and therefore $u(1) = 1$.

Let $\eta(\Delta,C) \in [0,1]$ be the utility from using a classifier $C$ for a dataset $\Delta$. We assume that the model with the highest utility is that which is most accurate, and thus utility could be based on the correct classification rate or other model performance metrics. We define cumulative utility across the model life-span as:
\begin{equation} \label{eq:cumulativeU}
    U = \sum_{t=1}^{T}\Bigl\{\eta(\Delta_t, C_{u(t)}) - \mu \, a_t\Bigl\},
\end{equation}
where $\mu \geq 0$ is a fixed cost of choosing `update'.

We wish to learn a policy $\pi$ which, at each time-point $t$, takes the currently available information and decides whether the to update the model. We assume the policy is probabilistic, such that we set $a_t = 1$ with a probability that is specified by the policy.

Let $\Pi$ be the space of policies and $U_{\pi}$ be the cumulative utility associated to a policy $\pi \in \Pi$. We aim to find a policy $\pi$ with favourable cumulative utility \eqref{eq:cumulativeU}, i.e.~considering the trade-off between updating more often (thereby increasing utility $\eta$) and incurring costs $\mu$. A good policy strikes a balance, making updates only when they would improve utility above the cost. 

In practice, as we do not know in advance how drift will occur, we cannot determine \emph{a-priori} the sequence of actions which maximises the cumulative utility. To bypass this, we propose a training environment, which simulates possible episodes of concept drift. We model the updating problem as an MDP, such that we can train a policy with a favourable cumulative utility in the environment using RL. We subsequently put the partially-developed policy in use, learning from its real world behaviour 
to improve the policy further.

\subsection*{Model Updating as a MDP}

We will now re-define the above setting as an MDP, that is a tuple $(S,\Omega,P,R,\gamma)$. Here, $S$ is the set of possible states. We let the current state $s_t \in S$ represent the information gained in a time window up until time $t$. The exact makeup of states is an arbitrary choice, but in this paper we include the following information:
\begin{itemize}
    \item Utility estimates $\eta(\cdot,\cdot)$ on a recent time window, where utility is set to the correct classification rate of the classifier on a given dataset.
    \item The utility of the classifier at the last update time $\eta(\Delta_{u(t-1)},C_{u(t-1)})$.
    \item Auxiliary information from evaluating the utility at the above time points; precision, recall and cross entropy loss of the classifier on the same time window as the utility is calculated on.
    \item Information about the model $C_{u(t-1)}$. In this paper we use a logistic regression model as the classifier, and thus include the current coefficients of said classifier in the state.
    \item Information on the time since the last update.
\end{itemize}

We take actions $a_t \in \Omega$ as described above. Further, $P$ is the unknown state-transition probability function which defines the probability of transitioning to any state $s' \in S$ from state $s \in S$ under action $a \in \Omega$.

The reward function, $r_t = R(s_t, a_t)$, gives the immediate reward received after taking action $a_t$ in the current state $s_t$. We define the reward function as:
\begin{equation}
    r_t = 
    \begin{cases} 
        \eta(\Delta_t, C_t) - \eta(\Delta_t, C_{u(t-1)}) - \rho & \text{if $a_t$ = 1}\\
        0 & \text{if $a_t$ = 0}\\
    \end{cases} \label{eq:reward}
\end{equation}

where $\eta(\Delta_t, C_t) - \eta(\Delta_t, C_{u(t-1)})$ is the change in accuracy from fitting a new classifier $C_t$ compared to the accuracy of the previous classifier $C_{u(t-1)}$ on the current dataset $\Delta_t$, and $\rho$ is a penalization term which indirectly determines the frequency of recommended updates, with a higher value for $\rho$ creating a higher barrier for choosing to update the classifier.

The choice of $r_t$ is dependent on the make up of states $s_t$, as well as the objective function that we wish to optimize. We expect there to be a dependency between our states $s_t$ and our rewards $r_t$. Specifically, changes in utility (or derived quantities) are likely to be predictive of the change in accuracy if a new classifier is fitted to the current data, $\eta(\Delta_t, C_t) - \eta(\Delta_t, C_{u(t-1)})$. We also expect this to be dependent on the time since the last classifier was fitted, i.e as $t$ moves further from $u(t-1)$, we generally expect $r_t$ to increase if an update has not been made in that time. We include the coefficients of the last classifier in $s_t$, as learned relationships with $r_t$ may depend on the classifier currently in use.

We can also say that the formulation of $r_t$ is roughly analogous to that within the sum in \eqref{eq:cumulativeU}, with the added benefit of being much easier to train as there is a greater difference between taking action $a_t = 1$ or $a_t = 0$. This formulation means that the penalization term $\rho$ is not identical to the fixed cost $\mu$ although it would be practical to set $\rho$ higher in settings where the fixed cost $\mu$ is known to be high. Furthermore, we can apply hyperparameter tuning to choose a value of $\rho$ which maximises \eqref{eq:cumulativeU} in simulated training data, this is discussed in the Experiments section. 

We specify the policy as a neural network with parameters $\xi \in \Xi$. We denote this neural network as $\pi_{\xi}(s)$ which we call the policy network, where $\xi$ indicates its parameters (weights and biases). The policy network takes as inputs states $s \in S$ and outputs probabilities of actions $a \in \Omega$. In RL, we search for the optimal parameters $\xi^* \in \Xi$ which maximise the expected return (discounted cumulative reward), that is:
\begin{equation}
    \pi_{\xi^*}(\cdot) = \arg\max_{\xi \in \Xi}\ \mathbb{E} \Bigl\{\sum_{k = 0}^{T}{\gamma^k r_{t+k}}\Bigl\},
\end{equation}
which we refer to as the optimal policy. We use the discount factor $\gamma$ as it would be impractical to optimize for expected reward over the whole time horizon $t,\dots,T$, so $\gamma$ allows us to control how far into the future we optimize. In this paper we choose a value of $\gamma = 0.8$ to optimize rewards in the more immediate future, while not totally ignoring longer term return. This is a hyperparameter which could also be tuned, but this is not the focus of the paper.

\section*{AMUSE}

Using the above MDP formulation, it may be possible to learn an updating policy in the real-world using RL methods. That is, you start with an observed dataset $\Delta_1$ and you have fitted a classifier $C_1$. As you observe incoming datasets $\Delta_2, \Delta_3, \dots$ you can create states, take actions and calculate rewards according to the above MDP. However, using traditional RL algorithms in this way would result in a policy that starts with little information, as future datasets are unknown. This may lead to a period in which we start with a highly non-optimal policy. This would not be ideal for settings in which there are significant downsides to an inaccurate model, or large costs of updating.

Suppose we have some knowledge about how the performance of $C_1$ may change. In particular, we know $C_1$, the current distribution of $(X,Y)$, and the likely rate and type of concept drift. It is hard to directly use this information to learn an updating policy. Thus, we wish to create a simulating environment to learn a policy which can be used to make the actions $a_t$ upon observing each future dataset $\Delta_t$. By using a simulating environment, we expect that this policy will perform better than random selection of actions. We can then update our policy upon exposure to real data. To do this we need to make assumptions about the data generating process and the drift generating process.

Let $m: \Theta \times \mathscr{X} \rightarrow [0,1]$ denote a parametric function defining the true data generating process, that is $m(x ; \theta) = Pr(Y=1|X=x)$ for some parameters $\theta \in \Theta$. Concept drift arises when $\theta$ is not constant, i.e.~the distribution $(Y|X)$ changes over time. We allow for change by denoting the parameters at time $t$ by $\theta_t$. Therefore, over the model life-span $T$ we can represent the drifting environment by a vector $\theta_{[T]} := (\theta_t)_{t\in [T]}$.

We assume that $\theta_{[T]}$ arises from a random process governed by a parameter $\phi$ taking values in some 
space $\Phi$. We will call this process $D(\phi)$. For instance, this may be a stationary Gaussian process with kernel parameterised by $\phi$. We assume that $\phi$ is time-invariant and determines the overall properties of drift; such as by governing how fast drift can occur or the range of values which parameters $\theta_t$ can take.

Our main observation is that, at any time $t$ if $\theta_{[t]}$, $\phi$ and the current model $C_{u(t)}$ are known, then it is possible to learn through simulation alone the optimal policy, that is the policy which maximises the expected cumulative utility. Here, the expectation is over the randomness in the data generating process and the drift generating process.

Thus, at time $t=1$ we can create a simulating environment by estimating the data generating process with a function $m(\cdot ; \hat{\theta}_{1})$, and by estimating the drift generating process with a function $D(\hat{\phi})$. Practically, these quantities can be estimated using expert opinion, and an initial exploratory analysis of $\Delta_1$. With these two quantities we can create Monte-Carlo simulations of possible drift paths, that is ultimately we can create simulated datasets $\Delta^*_2,\Delta^*_3,\dots,\Delta^*_{J}$ where $J$ is the length of the drift path. Here, it is not necessary for the length of the drift paths to equal the model life-span $T$. By taking actions in these paths (i.e.~updating the classifier or not), we can learn a policy using RL prior to observing future, potentially drifted datasets. Once real world datasets have been observed, we can update our policy using its observed performance, again using RL. 

Let $X_t$ be the matrix of observed covariates from $\Delta_t$, and $Y_t$ be the corresponding vector of observed responses. One episode of randomised drift can be simulated according to Algorithm \ref{alg:training} using our MDP formulation.

\begin{algorithm}[ht]
\caption{AMUSE: Simulating Environment}
\label{alg:training}
\textbf{Input: }\text{$X_1$, $Y_1$ from $\Delta_1$}\\ \text{Function $D(\hat{\phi})$ with known parameters $\hat{\phi}$,}\\
\text{Function $m(\cdot ; \hat{\theta}_1)$ with known parameters $\hat{\theta}_1$,}\\
\text{Policy $\pi$}\\
\text{Number of time steps $J$}\\
\begin{algorithmic}[1]
\STATE \text{Generate $\hat{\theta}_{[T]}$ using $D(\hat{\phi})$}
\STATE \text{Train an initial classifier $C_1$ ($a_1 = 1, r_1 = 0$)}
\STATE \text{Calculate:} $s_1$
\FOR{$t \in 2,...,J$}

    \STATE \text{Generate $Y^*_{t}$, $\Delta^*_{t}$ using $m(X_1 ; \hat{\theta}_{t})$}

    \STATE \textbf{Calculate:} Current state $s_t$
    
    \STATE Take action $a_t$ according to $\pi$

    \STATE \textbf{return} $r_t$

\ENDFOR
\end{algorithmic}
\end{algorithm}

\subsection*{Proximal Policy Optimization}

We use Proximal Policy Optimization (PPO)~\citep{Schulman2017ProximalAlgorithms} to train our policy network $\pi_{\xi}(\cdot)$. PPO is part of a family of RL methods called actor-critic methods, in which two neural networks are trained simultaneously. The policy network $\pi_{\xi}(\cdot)$ is also known as the actor. A separate neural network $V_{\iota}(\cdot)$ called the value network (critic) is concurrently trained, which takes as inputs states $s \in S$ and outputs the expected cumulative reward the agent will receive in state $s$. The purpose of the critic is to evaluate the performance of the policy being trained.

The initial training of the policy is detailed in Algorithm \ref{alg:PPOtraining}. This paper details a static implementation of PPO (the policy network is trained once in the simulating environment) as well as a dynamic implementation (the parameters of the policy network are updated upon exposure to real world states, actions, rewards). However the initial training is identical for both implementations. Note that in AMUSE, we only train one actor at a time, rather than multiple simultaneous actors as is common with PPO implementations. The agent collects data by interacting with the training environment (Algorithm \ref{alg:training}), and the policy is updated based on the collected experiences.

We leave a full description of the PPO algorithm as beyond the scope of this paper. However, we note the following important quantities:

\begin{itemize}
    \item The advantage functions $A_1, \dots, A_B$, where $B$ is the number of training steps collected before we update the parameters of the policy network. The advantages measure how much better an action is compared to the average action taken from a given state. This is based on the difference between the observed reward and the value network (which estimates the expected future reward). These estimates help the agent identify actions that led to higher-than-expected rewards, guiding the policy updates toward better decisions. 
    
    \item The clipped loss function $L^{\text{CLIP}}(\xi)$: This is the loss function used to update the policy network.  $L^{\text{CLIP}}(\xi)$ balances policy improvement and stability. The loss is optimized with respect to the policy parameters $\xi$ by clipping the probability ratio between the new policy and the old policy. This prevents excessively large updates that might degrade performance. The clipped objective ensures that the policy does not deviate too far from the previous one, thereby maintaining stable and incremental improvements.

    \item The mini-batch size $G$: This determines the number of experiences (state-action pairs) processed at each optimization step. Larger batch sizes result in increased computation time but more stable updates.

    \item The number of epochs ($K$): After collecting experience from the environment, PPO updates the policy by running multiple passes (epochs) over the data. \( K \) defines the number of times the collected data is used to optimize the policy. Larger values of \( K \) allow for more thorough optimization, but can increase the risk of over-fitting.

\end{itemize}

\begin{algorithm}[ht]
\caption{AMUSE: Proximal Policy Optimization}
\label{alg:PPOtraining}
\textbf{Input}: \text{Training Environment (Algorithm \ref{alg:training})}\\
\text{Number of training steps $B$}\\
\text{Mini-batch size $G$}\\
\text{Current policy $\pi_{\xi_{\text{old}}}(\cdot)$}\\
\begin{algorithmic}[1]
\FOR{iteration = 1, 2, ...}
    \STATE Run policy $\pi_{\xi_{\text{old}}}(\cdot)$ in Training Environment for $B$ time steps
    \STATE Compute advantage estimates $\hat{A}_1, ..., \hat{A}_{B}$
    \STATE Optimize $L^{\text{CLIP}}(\xi)$ wrt $\xi$ with $K$ epochs and mini-batch size $G \leq B$
    \STATE \text{Update Parameters}: $\xi_{\text{old}} \gets \xi$
\ENDFOR
\end{algorithmic}
\end{algorithm}

\subsection*{Implementing the Policy - Static vs Dynamic AMUSE}

At each time $t$, given the previous model $C_{u(t-1)}$ and current dataset $\Delta_t$, the state $s_t$ is computed, the policy $\pi_{\xi}(s_t)$ dictates action $a_t$ and receives reward $r_t$. We consider two possible implementations of AMUSE:

\begin{compactitem}
    \item[\textbf{Static}:] In which the policy network $\pi_{\xi}(\cdot)$ is trained on the simulating environment and subsequently is used to take actions on the real world datasets, but is never changed after the initial training.
    \item[\textbf{Dynamic}:] In which a policy network is first trained on the simulating environment and subsequently updated periodically after taking actions given states in the real world, and receiving rewards.
\end{compactitem}

In the static implementation of AMUSE, the policy network is solely trained within the simulating environment. On real world deployment, the policy can either be set as probabilistic (as in the training stage) or deterministic. In our experiments we use the probabilistic setting, that is we set $a_1$ according to the probabilities returned by the policy network. If deterministic, we choose $a_1 = 1$ if the probability returned by the policy network exceeds a pre-specified threshold.

The dynamic implementation of AMUSE includes further training of of the policy network. However, rather than using the simulating environment described by Algorithm \ref{alg:training}, we instead collect experiences on the real data batches as they arrive, resulting in Algorithm \ref{alg:PPOdynamic}. Here, it is natural to set the policy as probabilistic to ensure appropriate exploration.

For the dynamic version, the hyper-parameters (e.g.~$G$) used in the initial training stage can be set independently of the parameters used with further training the policy using real world data. For instance, it may be necessary to set a lower learning rate as we begin with a mostly trained policy in that case. Further, in many settings there are few discrete time points in the model life-span $T$, so parameters such as the number of number of training steps per update $B$ or the batch size $G$ may be lower in the real world environment compared to the initial training environment.

\begin{algorithm}[ht]
\caption{AMUSE: Dynamic Policy Updating}
\label{alg:PPOdynamic}
\textbf{Input}: \text{Data batches $\Delta_1,\dots,\Delta_T$}\\
\text{Pre-trained policy $\pi_{\xi_{old}}(\cdot)$}\\
\text{Number of training steps $B$}\\
\text{Mini-batch size $G$}\\
\begin{algorithmic}[1]
\STATE \text{Train an initial classifier $C_1$ on $\Delta_1$ ($a_1 = 1$)}
\STATE \text{Calculate:} $s_1$ \text{from $\Delta_1$}
\STATE \text{Set $r_1 = 0$}
\STATE Initialize counter $c = 0$
    \FOR{$t = 2,\dots,T$}
        \STATE \textbf{Calculate:} Current state $s_t$
        \STATE Take action $a_t$ according to the policy $\pi_{\xi}(s_t)$
        \STATE Observe reward $r_t$

        \STATE Store $(s_t, a_t, r_t)$
        \STATE Increment counter $c \gets c + 1$
        
        \IF{$c = B$}
            \STATE Compute advantage estimates $\hat{A}_1, ..., \hat{A}_{B}$
            \STATE Optimize $L^{\text{CLIP}}(\xi)$ wrt $\xi$ with $K$ epochs and mini-batch size $G \leq B$
            \STATE \text{Update Parameters}: $\xi_{\text{old}} \gets \xi$
            \STATE Reset counter $c \gets 0$
        \ENDIF
    \ENDFOR
\end{algorithmic}
\end{algorithm}

\section*{Experiments}

We assess the performance of AMUSE in two simulated concept drift settings; one in which we know the initial data generating process $m(\cdot ; \theta_1)$ and the drift process $D(\phi)$ (no model misspecification), and another in which both $m(\cdot ; \theta_1)$ and $D(\phi)$ are misspecified. An initial dataset $\Delta_1$ is generated, where the columns of $X_1$ are drawn from the standard normal distribution and $Y_1$ is drawn using $m(X_1 ; \theta_1)$. Then, drifted model parameters $\theta_{[500]}$ are simulated across 500 time points using random draws from $D(\phi)$. In turn, $\theta_{[500]}$ is used to generate future datasets in which drift has occurred ($\Delta_2,\dots,\Delta_{500}$), where at each $t$ we resample $X_t$ from a normal distribution. For all time points, datasets consist of 10000 independent samples. We repeat this process 50 times with different random seeds for the drift generating process. More details of the data generation can be found in the supplementary material.

For both the static and dynamic implementations of AMUSE, the policy network is trained using the simulating environment described in Algorithm \ref{alg:training}.

In the `no model misspecification' experiment, $m(\cdot;\cdot)$ is a defined as a logistic regression with $5$ features both when generating the data and as part of Algorithm \ref{alg:training} (defining $\hat{\theta}_{1}$ and $\hat{\phi}$ as the known values $\theta_1$ and $\phi$). Furthermore, $D(\phi)$ is a random walk of the parameters $\theta_t$, with a small possibility of sudden drift at each time step.

Meanwhile, in the `model misspecification' case, we also use a logistic regression specification for $m(\cdot;\cdot)$. We generate the data similarly to above, but with extra features including interaction terms, higher order terms and extra covariates. The drift generating process is also identical to above. However, when training AMUSE, we ignore the additional terms. Furthermore, we also misspecify the drift process $D(\cdot)$, underestimating the magnitude of drift and ignoring the possibility of sudden drift (see supplementary material). 

Hyperparameter tuning for the penalization term $\rho$ is conducted independently of comparative evaluation of updating policies on the basis of visual inspections of reward curves (plots of cumulative reward per episode in the training stage) for agents trained with each $\rho$ in $\{0.005, 0.01, 0.02, 0.03, 0.04\}$, as well as 10 extra runs of training environment where the cumulative utility is evaluated for different costs $\mu$, for each choice of $\rho$. We pick the value of $\rho$ which generally maximises cumulative utility across the most costs in the 10 runs, with a reward curve without too much noise. The discount factor $\gamma$ is set to $0.8$ on the basis of maximising shorter term rewards. All other hyperparameters are set as the defaults for the \texttt{Stable-Baselines3} python package \citep{Raffin2021Stable-Baselines3:Implementations}. See supplementary material.

Both static and dynamic agents are then allowed to make decisions about whether to update the classifier on datasets $\Delta_2,\Delta_3,...,\Delta_T$ for each randomised episode. For the classifier $C$ a logistic regression is used in all cases. The states are represented as a vector containing information including past model accuracies, precision, recall, coefficients and the time since the last update. Full model specification can be found in the supplementary material.

We compare AMUSE to methods chosen on the basis of frequency in the literature, adaptability to working with batches of data and ease of implementation. We compare the static and dynamic AMUSE approaches to the following updating strategies: 

\begin{itemize}
    \item \textbf{DDM}~\citep{Gama2004LearningDetection}: 
    if drift exceeds the `drift level' recommended by the algorithm, the model is re-trained on the current dataset $\Delta_t$. Default parameters from the paper are used for this method.
    \item \textbf{HDDM}~\citep{Frias-Blanco2015OnlineBounds}: similarly to DDM identifies a `drift level', which we use to inform updates should the test statistic exceed this level. Default parameters from the paper are also used for HDDM.
    \item \textbf{Random}: at each time step $t$ a decision of whether to re-train the model is made randomly, with probability chosen based on the average number of updates recommended by AMUSE for fairness.
    \item \textbf{Equally Spaced}: updates were spaced evenly across the life-span $T$ such that the final number of updates equalled the average number of updates recommended by AMUSE (rounded down).
    \item \textbf{Always update}: the initial classifier is subsequently re-trained for all time steps $t \in 2,\dots,T$ on the current dataset $\Delta_t$, and no other data.
    \item \textbf{Never update}: the initial classifier $C_1$ is never re-trained.
\end{itemize}
    
All strategies are assessed for the average cumulative utility \eqref{eq:cumulativeU} per episode where $\eta()$ is the correct classification rate of the model at each time-step, for different fixed values of the update penalty $\mu$.

\subsection*{No Model Misspecification}

Results can be seen in Table \ref{tab:table1}. We see that either dynamic or static AMUSE are the best performing method across all costs. In particular, both versions of AMUSE have average cumulative utilities with standard errors which do not overlap with the evenly spaced and random updating approaches. Given that these were chosen such that the number of updates matches those recommended by AMUSE, this indicates that AMUSE is able to learn a policy which is predictive of the cumulative reward above and beyond random chance. Furthermore, both versions of AMUSE also seem to marginally outperform both DDM and HDDM.

\begin{table*}[htbp]
\centering
\begin{tabular}{l|l|l|l|l|l}
    \multicolumn{1}{c|}{} & \multicolumn{5}{c}{\textbf{Update Cost, $\mu$}} \\ \cline{2-6}
    \textbf{Method} & \textbf{0.01} & \textbf{0.05} & \textbf{0.10} & \textbf{0.15} & \textbf{0.20} \\ \hline
    AMUSE (Dynamic) & 347.9 (1.07) & 346.7 (1.07) & 345.3 (1.08) & \textbf{343.9 (1.09)} & \textbf{342.5 (1.10)} \\
    AMUSE  (Static) & \textbf{348.2 (0.99)} & \textbf{346.9 (1.00)} & \textbf{345.4 (1.01)} & 343.8 (1.02) & 342.3 (1.03) \\
    DDM  & 342.6 (1.71) & 341.4 (1.70) & 340.0 (1.69) & 338.5 (1.68) & 337.0 (1.68) \\
    HDDM  & 346.6 (0.95) & 344.0 (0.98) & 340.8 (1.01) & 337.6 (1.04) & 334.4 (1.08) \\
    Random Updates  & 332.6 (1.80) & 331.4 (1.80) & 329.9 (1.79) & 328.4 (1.78) & 326.9 (1.77) \\
    Equally Spaced Updates  & 340.7 (1.12) & 339.5 (1.12) & 338.0 (1.12) & 336.6 (1.12) & 335.1 (1.12) \\
    Always Update  & 347.2 (0.92) & 327.2 (0.92) & 302.2 (0.92) & 277.2 (0.92) & 252.2 (0.92) \\
    Never Update & 263.8 (3.47) & 263.8 (3.47) & 263.8 (3.47) & 263.8 (3.47) & 263.8 (3.47)
\end{tabular}
\caption{Average cumulative utility over all 50 drift runs ($T=500$), for each method by update cost $\mu$ for the simulations without model misspecification. The values in parentheses are the standard errors.}
\label{tab:table1}
\end{table*}

\subsection*{Model Misspecification}

Results can be seen in Table \ref{tab:table2}. These show dynamic AMUSE is marginally the best performing method for all costs, although there is a great deal of uncertainty here. In this case, dynamic AMUSE outperforms static AMUSE across all costs $\mu$.

\begin{table*}[htbp]
\centering
\begin{tabular}{l|l|l|l|l|l}
    \multicolumn{1}{c|}{} & \multicolumn{5}{c}{\textbf{Update Cost, $\mu$}} \\ \cline{2-6}
    \textbf{Method} & \textbf{0.01} & \textbf{0.05} & \textbf{0.10} & \textbf{0.15} & \textbf{0.20} \\ \hline
    AMUSE (Dynamic) & \textbf{338.7 (1.05)} & \textbf{337.6 (1.05)} & \textbf{336.3 (1.04)} & \textbf{335.1 (1.04)} & \textbf{333.8 (1.04)} \\
    AMUSE (Static)  & 337.9 (1.04) & 336.9 (1.03) & 335.6 (1.03) & 334.3 (1.02) & 333.0 (1.02) \\
    DDM  & 334.6 (1.06) & 333.5 (1.07) & 332.2 (1.07) & 330.9 (1.08) & 329.6 (1.08) \\
    HDDM & 337.5 (0.99) & 335.2 (1.01) & 332.4 (1.03) & 329.5 (1.06) & 326.6 (1.09)\\
    Random Updates & 323.6 (1.38) & 322.5 (1.37) & 321.1 (1.36) & 319.7 (1.35) & 318.3 (1.34) \\
    Evenly Spaced Updates & 331.1 (1.20) & 330.1 (1.20) & 328.9 (1.20) & 327.7 (1.20) & 326.5 (1.20) \\
    Always Update & 338.4 (0.96) & 318.4 (0.96) & 293.4 (0.96) & 268.4 (0.96) & 243.4 (0.96) \\
    Never Update  & 263.8 (2.61) & 263.8 (2.61) & 263.8 (2.61) & 263.8 (2.61) & 263.8 (2.61) 
\end{tabular}
\caption{Average cumulative utility over all 50 drift runs ($T=500$), for each method by update cost $\mu$ for the simulations with model misspecification. The values in parentheses are the standard errors.}
\label{tab:table2}
\end{table*}

\subsection*{Summary of Experimental Findings}

Our key findings are:

\begin{itemize}
    \item AMUSE mostly outperforms common concept drift detection / updating methods DDM and HDDM on the simulated datasets. In particular, in the model misspecification case, the average recommended number of updates by dynamic AMUSE was 25.9, compared to 26.46 and 57.62 by DDM and HDDM respectively. Despite recommending fewer updates than HDDM, we are able to achieve comparable performance even when update costs are very low ($\mu = 0.01$). Likewise, despite recommending a similar amount of updates to DDM, we outperform it across all costs $\mu$
    \item In both the simulations AMUSE (static and dynamic) outperforms the random and equally spaced update approaches, despite these being calibrated to recommending roughly the same number of updates on average. This means that AMUSE is able to predict updates which will give a significant improvement in performance above and beyond random chance.
    \item  Dynamic updating of the policy network improves the performance of AMUSE above the static implementation in the misspecified experiment. This indicates that performance can be improved by updating the policy as more is learned about the previously misspecified drift generation process.
\end{itemize}

\section*{Conclusion}

The primary goal of this paper is to propose a method which recommends update times for a classifier, rather than merely detecting concept drift. To do this we have used an approach in which a policy is first learned in a simulating environment, and subsequently adapted to the real world concept drift. To our knowledge this is the first paper to consider such an approach, and as such we note the novelty, particularly of the use of a simulating environment in concept drift updating strategies.

The framework is generic, meaning it can be used on real world systems provided an initial dataset to train a classifier is available. Furthermore, the generic framework allows for extra elements to be included in the state representation that were not present in this work, such as distributional properties of the data. The training stage of the AMUSE algorithm allows for the specification of drift type to improve generalisability of the learned network. This is useful in real world scenarios where the anticipated drift is of unknown type.

We consider our approach a starting point to the development of further reinforcement learning methods to update models which use a simulating environment. Future work may include adaptation to continuous rather than batched data streams, exploration of more complex reward functions, states and penalties, and testing of this method on real world data. In this paper, we consider a stationary drift generating process; that is, the drift generating process does not itself change over time. In the real world this assumption is perhaps too strong, and implementations of AMUSE may need to consider re-running the simulating environment at certain time points with updated estimations of the drift generating process. AMUSE could also be extended to work with concept drift beyond just changes in the distribution of $(Y|X)$, for instance changes in the distribution of $(X)$ could be captured by including summary statistics of the datasets in the state space at each time point.

Finally, future work may consider how AMUSE performs over longer model-life spans $T$, particularly the dynamic model. In scenarios in which the model life-span is very large, dynamic AMUSE may be able to improve on static AMUSE further, and maintain a better policy for longer.

\section*{Declarations}

Louis Chislett acknowledges the receipt of studentship awards
from the Health Data Research UK-The Alan Turing Institute Wellcome PhD Programme in Health Data Science (Grant Ref: 218529/Z/19/Z). 

\subsection*{Conflict of interest} 
On behalf of all authors, the corresponding author states that there is no conflict of interest

\bibliographystyle{apalike}
\bibliography{references}

\end{document}